# LiDAR-3DGS:
# LiDAR Reinforced 3D Gaussian Splatting for Multimodal Radiance Field Rendering

Hansol Lim[1], Hanbeom Chang[1], Jongseong Brad Choi[1] and Chul Min Yeum[2]

Mechanical Engineering, State University of New York, Stony Brook, Stony Brook, NY 11794, USA
Civil Engineering Department, University of Waterloo, Waterloo, ON N2L 3G1, Canada

*Abstract*—In this paper, we explore the capabilities of multimodal inputs to 3D Gaussian Splatting (3DGS) based Radiance Field Rendering. We present LiDAR-3DGS, a novel method of reinforcing 3DGS inputs with LiDAR generated point clouds to significantly improve the accuracy and detail of 3D models. We demonstrate a systematic approach of LiDAR reinforcement to 3DGS to enable capturing of important features such as bolts, apertures, and other details that are often missed by image-based features alone. These details are crucial for engineering applications such as remote monitoring and maintenance. Without modifying the underlying 3DGS algorithm, we demonstrate that even a modest addition of LiDAR generated point cloud significantly enhances the perceptual quality of the models. At 30k iterations, the model generated by our method resulted in an increase of 7.064% in PSNR and 0.565% in SSIM, respectively. Since the LiDAR used in this research was a commonly used commercial-grade device, the improvements observed were modest and can be further enhanced with higher-grade LiDAR systems. Additionally, these improvements can be supplementary to other derivative works of Radiance Field Rendering and also provide a new insight for future LiDAR and computer vision integrated modeling.

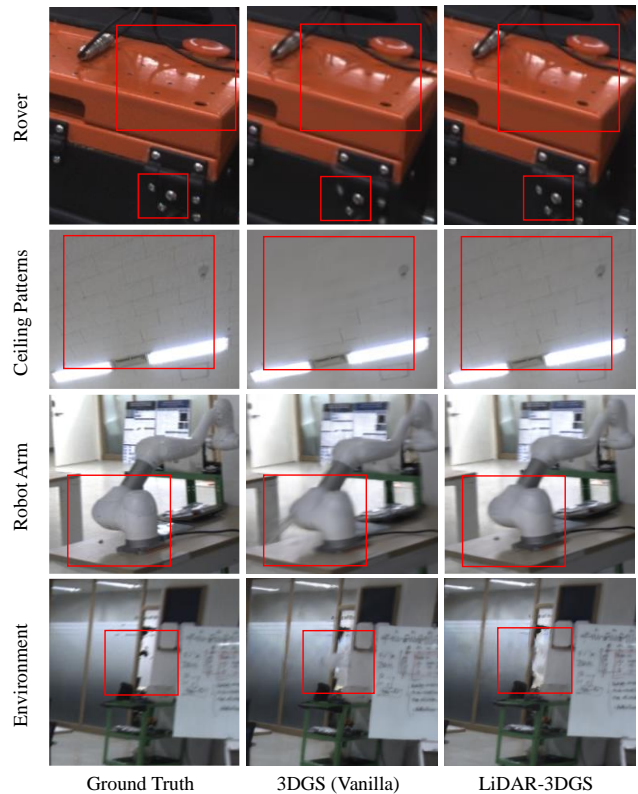

**Fig. 1.** Summary of qualitative assessment of various objects in LiDAR-3DGS model.

## I. INTRODUCTION

RADIANCE Field Rendering [1] emerged as a powerful technique for machine learning based 3D rendering. One of the notable projects among these is 3D Gaussian Splatting (3DGS). Originally introduced by Kerbl et al. in "3D Gaussian Splatting for Real-Time Radiance Fields Rendering (2023)", this method excels in generating high-quality 3D models from set of images. It works by creating point cloud through Structure-from-Motion (SfM) with images then converting the point cloud into a continuous volumetric field using Gaussian functions [2]. However, while 3DGS offers impressive results in rendering, its reliance on SfM point cloud alone limits its ability to capture fine geometric details, particularly in featureless or chromatically homogeneous environments, and objects [3], [4].

LiDAR (Light Detection and Ranging) technology, on the other hand, provides highly precise distance measurements, resulting in dense point clouds that capture the intricate geometries of physical objects [5]. Combining the strengths of LiDAR with 3DGS presents an opportunity to enhance the accuracy and detail of 3D models significantly. This integration is particularly valuable in applications requiring high-fidelity representations, such as engineering, remote monitoring, and infrastructure maintenance [6], [7].

This paper presents LiDAR-3DGS, a systematic approach

This work supported by the National Research Foundation of Korea 14 CHOI et al. (NRF) grant funded by the Korea government (MSIT) (No. 2022M1A3C2085237). *(Corresponding author: Jongseong Brad Choi).*

Hansol Lim, Hanbeom Chang, and Jongseong Brad Choi are with Mechanical Engineering, State University of New York, Stony Brook, Stony Brook, NY 11794, USA (e-mail: hansol.lim@stonybrook.edu; hanbeom.chang@stonybrook.edu; jongseong.ch oi@stonybrook.edu)

Chul Min Yeum is with the Civil Engineering Department, University of Waterloo, Waterloo, ON N2L 3G1, Canada (e-mail: cmyeum@uwaterloo.ca)

to reinforce LiDAR point cloud data to 3DGS. By leveraging the complementary strengths of vision-based and LiDAR data, LiDAR-3DGS captures intricate features such as bolts, holes, and microcracks that are often missed by SfM alone. Our approach demonstrates that even a modest increase in LiDAR density can substantially enhance the perceptual quality of the models without modifying the underlying 3DGS algorithm.

The contributions of this paper are as follows:
1) We propose a systematic approach for integrating LiDAR data into 3DGS.
2) We propose how the multimodal input can improve 3DGS modeling of featureless or chromatically homogenous environments and objects.
3) We experimentally validate the improvements by demonstrating the method's ability of capturing fine details supported by widely used performance metrics.

The remainder of this paper is organized as follows: Section 2 reviews related literature, Section 3 details the proposed methodology for reinforcing 3DGS with LiDAR data. In Section 4, we present our experimental setup and validation results, demonstrating the effectiveness of our approach. Finally, Section 5 concludes the paper and discusses some of the limitations of the study.

## II. LITERATURE REVIEW

### A. 3D Gaussian Splatting

3D Gaussian Splatting, introduced by Kerbl et al. (2023), is a follow-up breakthrough of the revolutionary NeRF (Neural Radiance Fields). This technique excels in generating photorealistic 3D models from point cloud data, offering advantages in computational efficiency and rendering quality. The original work demonstrated the potential of 3DGS in various applications, including real-time radiance fields rendering.

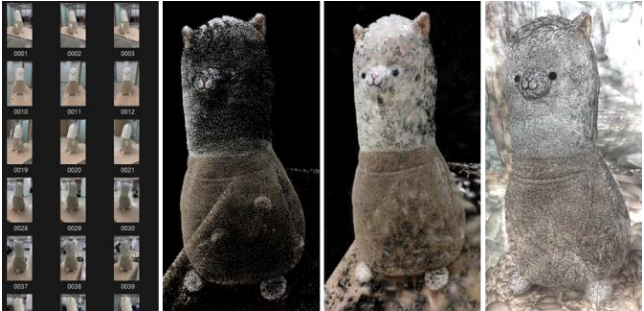

**Fig. 2.** 3D Gaussian Splatting process demonstration.

Initially, set of images are captured from a static scene. The corresponding cameras are calibrated using SfM, and produces a sparse point cloud. From the sparse point cloud, set of 3D Gaussians is created. Each 3D Gaussian is defined by four learnable parameters: Gaussian center $x_i \in \mathbb{R}^3$, 3D covariance matrix made from scale vector $s_i \in \mathbb{R}^3$, rotation vector $q_i \in \mathbb{R}^4$, its opacity $\alpha_i \in \mathbb{R}$, and its spherical harmonics $SH_i \in \mathbb{R}^{48}$. The Gaussian function is defined as:

$$G(x; \mu, \Sigma) = \frac{1}{(2\pi)^{\frac{3}{2}}|\Sigma|^{\frac{1}{2}}} exp\left(-\frac{1}{2}(x-\mu)^T \Sigma^{-1}(x-\mu)\right) \quad (1)$$

Where $G(x; \mu, \Sigma)$ represents the Gaussian function at a point x in 3D space, μ denotes the mean vector, which defines the center of the Gaussian splat in 3D space. The covariance matrix Σ describes the spread and orientation of the Gaussian function, indicating how the splat extends in different directions. The determinant of the covariance matrix, $|\Sigma|$, influences the normalization factor, ensuring the Gaussian integrates correctly over the entire space.

The 3D Gaussians are optimized through successive iterations. The goal of these iterations is to minimize the difference between the rendered scene and the input images. This difference is quantified using a loss function provided by the original paper. It is proposed to represent the loss between the model and the data. It combines two metrics: the L1 loss and the structural similarity index.

$$L = (1-\lambda)L_1 + \lambda L_{D-SSIM} \quad (2)$$

$$L_1 = \frac{1}{N}\left(\sum_{i=1}^{N} \|I_{observed}(x_i, y_i) - I_{rendered}(x_i, y_i)\|^2\right) \quad (3)$$

$$SSIM(x, y) = \frac{(2\mu_x \mu_y + C_1)(2\sigma_{xy} + C_2)}{(\mu_x^2 + \mu_y^2 + C_1)(\sigma_x^2 + \sigma_y^2 + C_2)} \quad (4)$$

Where $L_1$ is the mean absolute error between the observed and rendered intensities. $I_{observed}(x_i, y_i)$ is the observed intensity at pixel $(x_i, y_i)$, $I_{rendered}(x_i, y_i)$ is the rendered intensity, and $N$ is the total number of pixel, $L_{D-SSIM}$ is the structural dissimilarity index $L_{D-SSIM} = (1 - SSIM)$. $\lambda$ is a weighing factor and is set to 0.2 according to the original paper.

To compare the rendered scene with the input images, the 3D scene is rasterized into 2D. During the rasterization process, the intensity at each pixel is computed by summing the contributions of all Gaussian splats. The intensity calculation at each pixel $(x_i, y_i)$ is as follows:

$$I(x_i, y_i) = \sum_k G_k(x_i, y_i) \quad (5)$$

Where $I(x_i, y_i)$ represents the intensity or color at pixel coordinates $(x_i, y_i)$, and $G_k(x_i, y_i)$ is the value of the $k$-th Gaussian splat. Complete deviations of gradient descent and rasterization process used in 3D Gaussian Splatting is in appendix section A and C respectively in the original paper.

### B. LiDAR data and Image data Integration

The integration of LiDAR with image data addresses the limitations of each individual technology, combining the precise geometric data from LiDAR with the rich color and texture information from images. There exists previous works that proved the improved 3D reconstruction capabilities of fusing LiDAR and Camera images such as that by Zhen et al. (2019), demonstrate the effectiveness of LiDAR and SfM point cloud fusion in generating 3D reconstruction. The fused data sets provide much denser spatial and visual information, resulting in improved 3D model mesh [8]. With the rise of computer vision and machine learning, this multimodal integration has been explored even further. Recent advancements have seen the application of deep learning techniques to fuse LiDAR and visual data. Tao, et al. (2023)

proposed a method of novel LiDAR view synthesis through NeRF, significantly enhancing LiDAR mapping capabilities [9]. This integration has proven particularly beneficial in applications requiring high-fidelity representations, such as LiDAR based autonomous driving [10].

GauU-Scene, introduced by Xiong et al. (2024), presents a large-scale scene reconstruction by combining LiDAR data with 3DGS. Their LiDAR-fused 3DGS model evaluation includes comparisons across various novel viewpoints and compares these results with original 3DGS [11]. Their comparison highlighted some differences, underscoring the importance of combining multimodal information for accurate 3D scene reconstruction. Through visual comparisons, it has demonstrated that the integration of LiDAR with Gaussian Splatting boosts the accuracy of 3DGS results. However, their results did not show a significant quantitative difference. This emphasizes the need for more investigations to LiDAR-3DGS integration and a systematic benchmarking of quality metrics.

### III. METHODOLOGY

#### A. Overview

The methodology is divided into several key phases: raw data collection, data sampling, point cloud alignment, and model training.
1) Raw Data Collection: We used a custom setup to calibrate and collect color-mapped LiDAR data.
2) LiDAR Data Sampling: To efficiently integrate the LiDAR point cloud with the SfM point cloud, we developed "ChromaFilter" algorithm. This algorithm reduces the density of the LiDAR point cloud by retaining essential structural features based on their color profiles, ensuring that only the most relevant points are used in the training process.
3) Camera Data Sampling: We also optimized the camera data sampling process by implementing an overlap-based sampling technique. This ensures that the images used for SfM have meaningful overlaps.
4) Point Cloud Alignment: Aligning the LiDAR and SfM point clouds into a same coordinate system is critical for 3DGS process. This is achieved through the Iterative Closest Points (ICP) algorithm.
5) Model Training: Finally, the aligned point clouds are used to train the 3D Gaussian Splatting model. We conducted experiments to determine the optimal parameters for different LiDAR data density and increment steps for spherical harmonics levels.

#### B. Raw Data Collection

To acquire a precise and robust color-mapped LiDAR dataset for this study, a custom module was developed [12]. One of the main challenges in LiDAR mapping is dealing with sensing inaccuracies. Duplicated walls or lost path tracking may occur if the LiDAR and RGB camera are not tightly coupled [13], [14], [15]. This problem can be solved by the extrinsic calibration of the sensors' coordinate systems and the synchronization of sensor data [16], [17] within the ROS (Robot Operating System) framework. The research utilizes ROS Package "camera_calibration" [18] and "direct_visual_lidar_calibration" package by koide3 [19], available on GitHub.

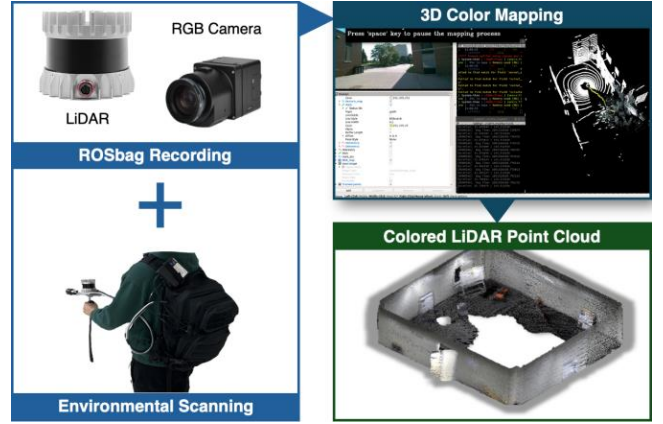

**Fig. 3.** Raw data collection process.

We used the Ouster OS0-32 LiDAR system [20], which offers a 360° horizontal and a 90° vertical field of view, divided into 32 channels. It has a resolution of 1024 points per horizontal scan, with a range of 0.3m to 35m. The sensor's rotation rate can be configured between 10 and 20Hz, providing up to 655,360 points per second at a data rate of 66Mbps. For the RGB camera, we employed the FLIR Blackfly S, which features a resolution of 1440x1080 pixels and a maximum frame rate of 60 FPS. With an additional lens, the camera's field of view is approximately 70°, and it uses the ½.9-inch Sony IMX273 CMOS sensor.

#### C. LiDAR Data Sampling: ChromaFilter

A key step in the methodology is subsampling LiDAR point cloud suitable to be fused with the sparse SfM point cloud. The objective was to streamline dense LiDAR point cloud, reducing it to a sparser form that retains essential structural features to be used in training 3D Gaussian Splats. We devised *ChromaFilter* [21] an algorithm to utilize RGB data of the point cloud. The characteristic that non-essential areas, such as walls and floors, were often homogenous in color while important features present a more diverse color range was exploited. The algorithm starts with a statistical outlier removal to eliminate noise from the LiDAR point cloud, ensuring that only relevant data are subjected to subsampling. The color distribution of the point cloud was then mapped by scaling the RGB values to the standard 255 scale, converting these to integer tuples to group points by their color profiles.

$$d(p_i) = \frac{1}{k} \sum_{p_j \in N_k(p_i)} |p_i - p_j| \quad (10)$$

$$P_{RGB}(p_i) = [R_i, G_i, B_i] \quad (11)$$

$$P_{random} = Random(P_i, n) \quad (12)$$

For each group of points sharing the same color, $P_{RGB}(p_i)$, a controlled reduction was implemented. If the group's size exceeded the set maximum points per color value, n, a random sample from this group up to the set limit was taken. This ensured a balanced representation across the spectrum, effectively thinning out areas of lesser importance while preserving features in regions that are more significant. The result was a sparse LiDAR point cloud with distinct features preserved and its density selectively reduced in a manner that aligns with the requirements of 3DGS. The density was determined based on choosing the right value for $n$, maximum points per color value. This parameter is tested later in the chapter to control the level of detail included in the model.

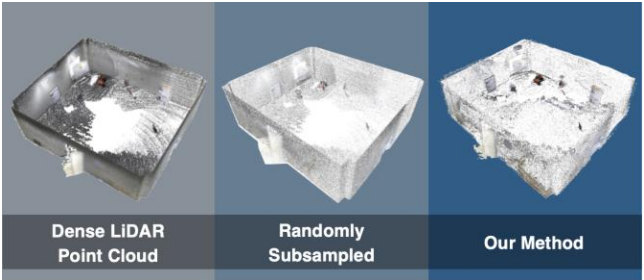

**Fig. 4.** Comparison between random sampling and our method.

*D. Camera Data Sampling: Overlap-based Sampling*

The raw camera data were recorded at 60fps. Given the high volume of images, a critical step was to sample these images effectively for SfM, ensuring meaningful overlap for 3D reconstruction. To achieve this, an algorithm was developed to estimate image overlap and select frames that maintain about 80% overlap with adjacent images.

The homography matrix between adjacent images was first computed using the ORB (Oriented FAST and Rotated BRIEF) [23] feature detector for its efficiency in identifying and matching robust features. ORB combines FAST [24] feature detection with a rotation-aware BRIEF [25] descriptor, offering a balance between performance and computational efficiency. ORB detects key points based on the intensity of a pixel compared to its surrounding neighborhood:

$$C(p) = \begin{cases} 1, & if \sum_{i=1}^{16}[I(p_i) > I(p) + t] \geq n \\ 1, & if \sum_{i=1}^{16}[I(p_i) < I(p) - t] \geq n \\ 0, & otherwise \end{cases} \quad (13)$$

The binary descriptor is constructed using a series of binary tests, $\tau(p,q)$, to compare the intensity of one pixel to other.

$$\tau(p,q) = \begin{cases} 1, & if\ I(p) < I(q) \\ 0, & otherwise \end{cases} \quad (14)$$

Following key point detection, it then calculates moments to determine the centroid and orientation of the key points to ensure the feature rotation invariance.

$$m_{pq} = \sum_{x,y} x^p y^q I(x,y) \quad (15)$$
$$C = \left(\frac{m_{10}}{m_{00}}, \frac{m_{01}}{m_{00}}\right) \quad (16)$$
$$\theta = atan2(m_{01}, m_{10}) \quad (17)$$

The Brute Force Matcher with Hamming distance was then applied, alongside the RANSAC algorithm to refine the homography matrix which directly affects accuracy of matched features:

$$H = arg\ min_H \sum_{i=1}^{n} \|p'_i - Hp_i\|^2 \quad (18)$$
$$H^* = RANSAC(\{p_i, p'_i\}) \quad (19)$$

With $H^*$ determined, the coordinates of each point in one image can be transformed to align with its match in the adjacent image.

$$X' = H^* X \quad (20)$$

Where the $H^*$ and the new pixel coordinate transformations, $X'$ is the point location in 3D, $X$ is the pixel location in 2D. The expression can be can be expanded as:

$$\begin{pmatrix} x' \\ y' \\ 1 \end{pmatrix} = \begin{pmatrix} h^*_{11} & h^*_{12} & h^*_{13} \\ h^*_{21} & h^*_{22} & h^*_{23} \\ h^*_{31} & h^*_{32} & 1 \end{pmatrix} \begin{pmatrix} x \\ y \\ 1 \end{pmatrix} \quad (21)$$

$$x' = \frac{h^*_{11}x + h^*_{12}y + h^*_{13}}{h^*_{31}x + h^*_{32}y + 1} \quad (22)$$

$$y' = \frac{h^*_{21}x + h^*_{22}y + h^*_{23}}{h^*_{31}x + h^*_{32}y + 1} \quad (23)$$

Upon successfully warping the image using $H^*$, the next step involves preparing both the warped image and the target image for overlap calculation. This involves converting both images to grayscale to simplify the data for the subsequent stages of processing.

The overlap between the warped and adjacent images is then calculated. To do this, both images are converted into binary form, where the presence or absence of features is distinctly marked. A bitwise AND operation is performed on these binary images to identify the overlapping areas and is quantified by the following formula:

$$OL\% = \frac{\sum_{i,j} I_{warped,adj}(i,j) \cap I_{warped,}(i,j)}{\sum_{i,j} I_{warped}(i,j)} * 100 \quad (24)$$

This procedure is repeated with subsequent images until the 80% overlap is achieved, selecting the image, and continuing until all the images are exhausted.

To address lens distortion inherent in the camera images, COLMAP simple_pinhole function [25] can be used to correct both radial and tangential distortion, transforming distorted pixel locations $(x_d, y_d)$ to their corrected positions $(x_u, y_u)$ as follows:

$$\begin{aligned} x_u &= x_d(1 + k_1 r^2 + k_2 r^4 + k_3 r^6) \\ &\quad + 2p_1 x_d y_d + p_2(r^2 + 2x_d^2) \end{aligned} \quad (25)$$

$$\begin{aligned} y_u &= y_d(1 + k_1 r^2 + k_2 r^4 + k_3 r^6) \\ &\quad + 2p_1 x_d y_d + p_2(r^2 + 2x_d^2) \end{aligned} \quad (26)$$

Where $r^2 = x_d^2 + y_d^2$ indicates the squared distance from the image center, and $k_1, k_2, k_3, p_1, p_2$ are the radial and tangential distortion coefficients, respectively. By removing distortion in each image in the dataset, the data geometry is corrected prior to subsampling for SfM processing. Then, these images were processed with SfM to create a point cloud.

*E. Point Cloud Alignment*

To be used in 3DGS training, the LiDAR point cloud needs to be aligned to the same coordinate system as SfM point cloud. We used CloudCompare to manually scale and rotate the LiDAR point cloud to match SfM point cloud's coordinates. With the coarse alignment done, the ICP (Iterative Closest Points) algorithm [26] was applied to minimize the error between the point clouds:

$$q_i = argmin_{q \in Q} \|p_i - q\|^2 \quad (27)$$

$$E(R, t) = \sum_{i=1}^{n} \|q_i - (Rp_i + t)\|^2 \quad (28)$$

$$(R^*, t^*) = argmin_{R,t} E(R, t) \quad (29)$$

The ICP algorithm works by finding corresponding points between point clouds and minimizing the error of these correspondences. This iterative process refines the alignment in two phases. The first phase aims to align major structures with a 90% match. Then, a second iterative loop is applied to aim for a 99% match.

$$P' = Iterate(RP + t), \quad threshold = [90\% \rightarrow 99\%] \quad (30)$$

Upon completion of point cloud fusion, the merged point cloud data is saved as a single .PLY file. This file serves as the final processed input for the 3D Gaussian Splat training.

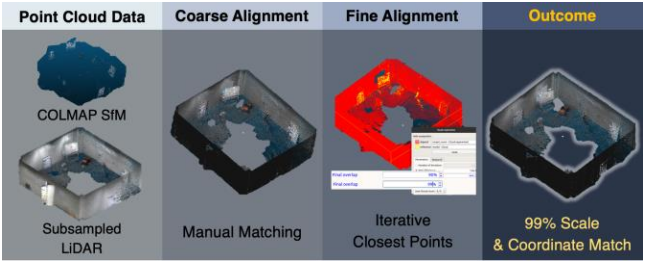

**Fig. 5.** Point cloud alignment process.

*F. Model Training*

The integration of LiDAR point cloud data into the SfM point cloud data significantly densifies the dataset for training 3D Gaussian splats. This led to the hypothesis that controlling the increment iterations for spherical harmonics levels can maximize the use of LiDAR point cloud in 3D Gaussian Splat training. Initially, we anticipated that lower spherical harmonics levels will exhibit more sensitivity to the denser LiDAR point cloud and can increase modeling accuracy for prominent features such as walls and floors. To test this hypothesis, an experiment was set up to observe how varying timings of spherical harmonics level increment affect model quality.

In addition, we also experimented across range of varying LiDAR point cloud densities produced from ChromaFilter to see if more LiDAR point cloud aids in producing a better 3D model. If that is the case, we wanted to see the trade-off between model quality and increase in computational time.

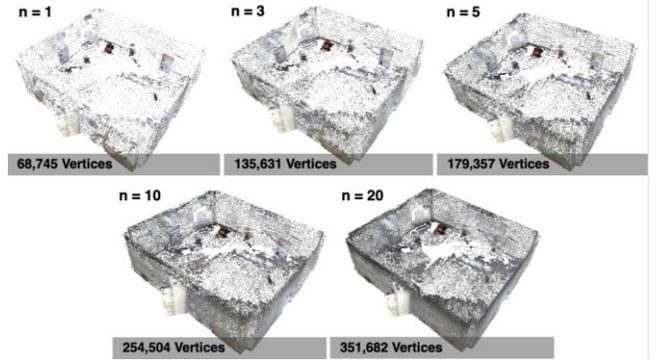

**Fig. 6.** LiDAR point clouds of different densities

*G. Model Performance Assessment*

The impact on two critical quality metrics was measured: Peak Signal-to-Noise Ratio (PSNR) and Structural Similarity Index Measure (SSIM) (5). PSNR metric is used to measure the quality of a reconstructed image or 3D model by comparing it to a reference image or model. It is useful for quantifying the amount of noise or error in the reconstructed 3D model. Higher PSNR values indicate lower error and higher fidelity to the reference model. It also serves as a reference to geometric precision as it measures the overall error of the model [27].

$$PSNR = 10 \, log_{10} \left( \frac{MAX_I^2}{MSE} \right) \quad (31)$$

Where MAX is the maximum pixel value of the image, MSE is the mean square error between the corresponding pixels in observed and rendered images.

On the other hand, SSIM is used to measure the perceived quality of an image or 3D model based on the structural similarity between the reference and the reconstructed model. It takes account of changes in textural information, luminance, and contrast. SSIM is more aligned with human visual perception and is particularly useful when we are assessing the perceived quality of the details and textures present in the 3D model. These metrics provide a quantitative insight into the reconstruction's accuracy and the visual quality of the resulting 3DGS model.

IV. VALIDATION

*A. Experiment Set-up*

The system used in the experiment is as follows:

## TABLE I
## SYSTEM SPECIFICATIONS

| CPU | Intel i7-13000KF |
|---|---|
| M/B | ASRock Z790 PG Lightning |
| GPU | NVIDIA RTX 4080 16GB |
| RAM | Crucial DDR5 64GB |
| SSD | SAMSUNG 990 Pro 1TB |
| **Camera: FLIR Blackfly S Specifications** | |
| Resolution | 1440 x 1080 |
| Frame Rate | 60 FPS |
| Field of View | 70° |
| Camera Sensor | SONY IMX273 |
| CMOS Sensor Size | 9 ½" |
| **LiDAR: Ouster OS0-32 Specifications** | |
| Horizontal Scan | 360° |
| Vertical Scan | 90° |
| Maximum Range | 35m |
| Resolution | 1024 x 24 Hz |
| Data Transfer | 655,360 points/s |

Experimental data were collected from the Research Lab, which encompasses approximately 300 square meters. The collected data included 207MB of LiDAR point cloud data along with 10,515 camera images. Initially, distortion was removed from the sampled camera data, followed by performing SfM to create an initial point cloud from the camera images.

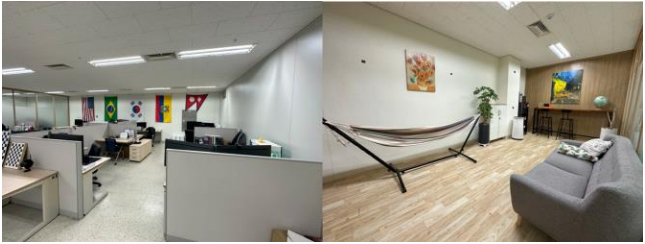

**Fig. 7.** Lab space used for dataset generation

After sampling the camera data with 85% overlaps, 245 images were selected. These images were undistorted and downscaled to 1600 x 1102 pixels. During the feature matching process, 22 images remained unmatched in the SfM and were therefore excluded from being used as inputs. Below is the summary of the camera data before and after the sampling, undistortion, and downscaling

## TABLE II
## CAMERA DATA SUMMARY

| | Camera Data summary | | | |
|---|---|---|---|---|
| Type | Raw Data | Sampled | Undistorted | Downscaled (Final input) |
| # of images | 10,516 | 245 | 223 (22 images were unmatched) | 223 |
| Resolution | 1440 x 1080 | 1440 x 1080 | 1714 x 1181 | 1600 x 1102 |
| Size | 22.3 GB | 532 MB | 221 MB | 209 MB |

For the LiDAR data, a point cloud comprising 5,241,214 total vertices was collected. LiDAR densities of n = 1, n = 5, n = 10, and n = 20 were extracted.

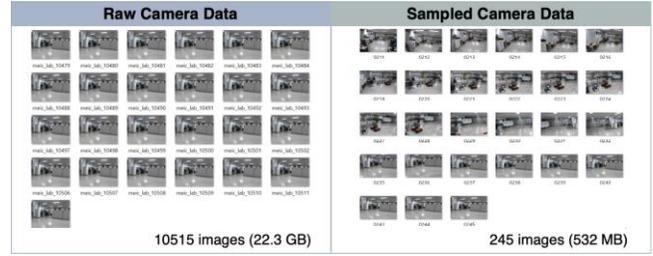

**Fig. 8.** Processed and sampled camera images

## TABLE III
## POINT CLOUD DATA SUMMARY

| Point Cloud Data summary | | | | | | |
|---|---|---|---|---|---|---|
| Density | Raw Data | SfM | n = 1 | n = 5 | n = 10 | n = 20 |
| # of vertices | 5,241,214 | 36,423 | 267,888 | 737,731 | 1,063,365 | 1,490,207 |
| Size | 191 MB | 1.38 MB | 6.89 MB | 18.9 MB | 27.3 MB | 38.3 MB |

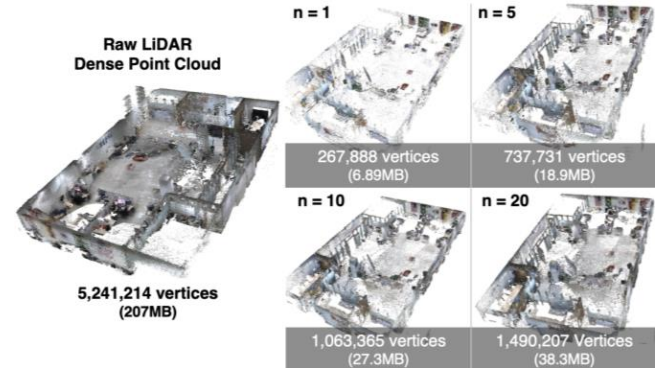

**Fig. 9.** Processed point clouds of different density.

### B. Experimental Results

The tables display PSNR and SSIM values at 30k iterations across different SH level increments and LiDAR densities. Yellow indicates best result; orange indicates second best result.

## TABLE IV
## PSNR AT 30K ITERATIONS

| PSNR at 30k iterations (Higher is better) | | | | | |
|---|---|---|---|---|---|
| SH/density | Vanilla | n = 1 | n = 5 | n = 10 | n = 20 |
| 500 | 33.1198929 | 34.4726955 | 34.6057114 | 34.7398285 | 34.6850189 |
| 750 | 33.1795853 | 34.8515900 | 34.7986748 | 34.6301132 | 34.6075859 |
| 1000 | 32.7039410 | 34.4277626 | 34.7580009 | 34.7032631 | 34.9383484 |
| 1250 | 32.4574493 | 34.4285957 | 34.3867077 | 34.8496956 | 34.4017418 |
| 1500 | 32.1640087 | 34.6672348 | 34.6379402 | 34.9688881 | 35.1166679 |
| 2000 | 31.6260307 | 34.5733765 | 34.8653717 | 34.9487389 | 34.7843159 |
| 2500 | 32.0547962 | 34.8537773 | 34.2112888 | 34.7029968 | 34.4314453 |
| 3000 | 33.0526009 | 34.4598022 | 34.4421188 | 34.8102409 | 34.3926712 |
| 4000 | 31.7322701 | 34.0821747 | 34.4910675 | 34.3687229 | 34.3899315 |
| Average | 32.4545083 | 34.5352233 | 34.5774313 | 34.7469431 | 34.6386363 |

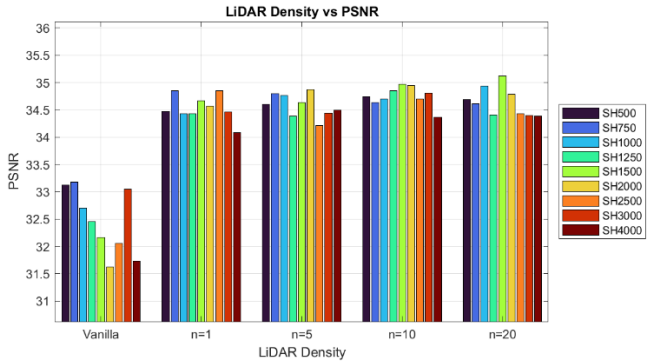

**Fig. 10.** PSNR at 30k iterations.

TABLE V
SSIM AT 30K ITERATIONS

| SSIM at 30k iterations (Higher is better) | | | | | |
|---|---|---|---|---|---|
| SH/density | Vanilla | n = 1 | n = 5 | n = 10 | n = 20 |
| 500 | 0.9528430 | 0.9567515 | **0.9575180** | 0.9580046 | **0.9582846** |
| 750 | **0.9533212** | 0.9567318 | 0.9571464 | 0.9581760 | 0.9577327 |
| 1000 | **0.9537780** | 0.9565855 | 0.9572202 | **0.9581863** | 0.9579604 |
| 1250 | 0.9519798 | **0.9569154** | 0.9574067 | **0.9581991** | 0.9577163 |
| 1500 | 0.9523735 | 0.9568077 | **0.9574783** | 0.9580824 | 0.9577348 |
| 2000 | 0.9527618 | 0.9565358 | 0.9572777 | 0.9581124 | 0.9579788 |
| 2500 | 0.9525703 | **0.9568613** | 0.9573013 | 0.9581817 | 0.9578961 |
| 3000 | 0.9525055 | 0.9562702 | 0.9574659 | 0.9579670 | 0.9579121 |
| 4000 | 0.9518756 | 0.9560952 | 0.9569930 | 0.9576808 | **0.9580220** |
| Average | 0.9526676 | 0.9566172 | 0.9573120 | 0.9580656 | 0.9579153 |

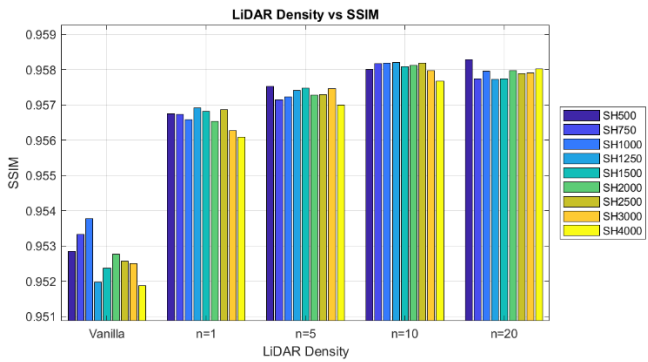

**Fig. 11.** SSIM at 30k iterations.

TABLE VI
LOSS AT 30K ITERATIONS

| Loss at 30k iterations (Lower is better) | | | | | |
|---|---|---|---|---|---|
| SH/density | Vanilla | n = 1 | n = 5 | n = 10 | n = 20 |
| 500 | 0.0180154 | 0.0174833 | 0.0169193 | 0.0169872 | **0.0162753** |
| 750 | **0.0178892** | 0.0174257 | 0.0170673 | 0.0170043 | 0.0165984 |
| 1000 | 0.0184445 | 0.0172770 | **0.0167617** | 0.0169567 | 0.0169635 |
| 1250 | 0.0185299 | 0.0173953 | 0.0172882 | 0.0172225 | 0.0171542 |
| 1500 | 0.0181508 | 0.0171975 | 0.0168628 | **0.0167531** | 0.0170607 |
| 2000 | 0.0189201 | 0.0173894 | **0.0168113** | 0.0169461 | 0.0171144 |
| 2500 | **0.0179494** | **0.0171936** | 0.0171266 | 0.0171686 | **0.0165324** |
| 3000 | 0.0185559 | **0.0169286** | 0.0170089 | 0.0171637 | 0.0166029 |
| 4000 | 0.0187975 | 0.0176292 | 0.0170963 | **0.0168935** | 0.0169810 |
| Average | 0.0184603 | 0.0173244 | 0.0169936 | 0.0170106 | 0.0168092 |

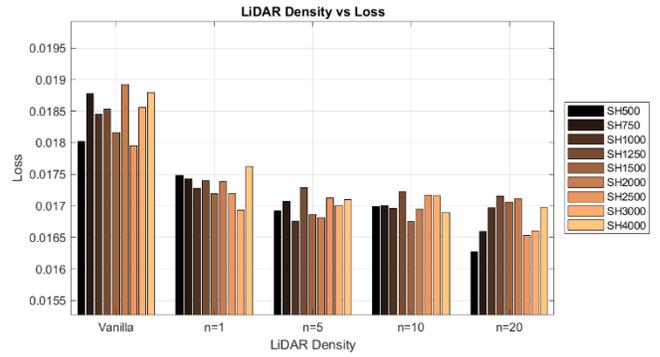

**Fig. 12.** Loss at 30k iterations.

TABLE VII
TIME TAKEN AT 30K ITERATIONS

| Time taken at 30k iterations at SH1000 (Lower is better) | | | | |
|---|---|---|---|---|
| Vanilla | n = 1 | n = 5 | n = 10 | n = 20 |
| 19m 51s | 20m 30s | 22m 17s | 23m 3s | 23m 44s |

*C. Qualitative Evaluation*

For qualitative analysis, various objects and environments were selected in the resulting 3D model. Objects were compared at the same camera angle as the ground truth side-by-side. At every corner, the LiDAR-3DGS model proved to be superior to the vanilla 3DGS model. Objects were more clearly segregated with less errors. The LiDAR point cloud help mold better Gaussian splats around the objects, resulting in a more accurate and detailed representation.

We compared n=20, SH1500 configuration LiDAR-3DGS model against the vanilla 3DGS model.

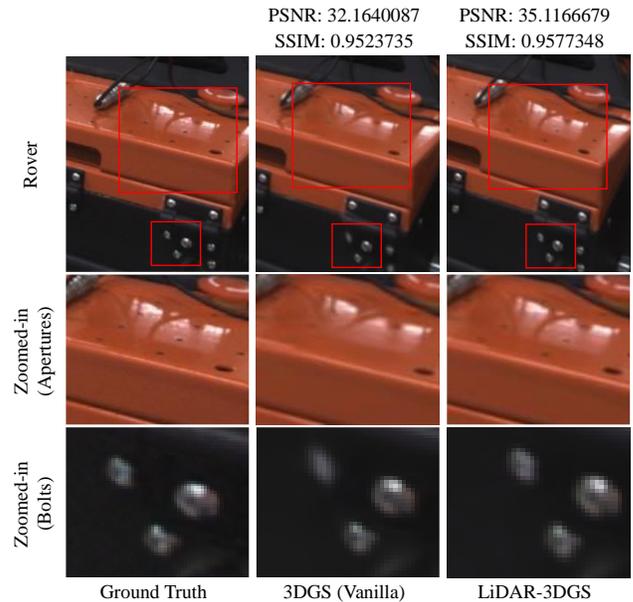

**Fig. 13.** Qualitative assessment of rover model.

The vanilla 3DGS model failed to reflect the apertures and bolts present on the rover's hood, which are all crucial details. In contrast, the LiDAR-3DGS model accurately captured these details, demonstrating the enhanced precision and detail achievable with the multimodal inputs in 3DGS. The denser point cloud from LiDAR supports the SfM feature points, enabling a more accurate and detailed model.

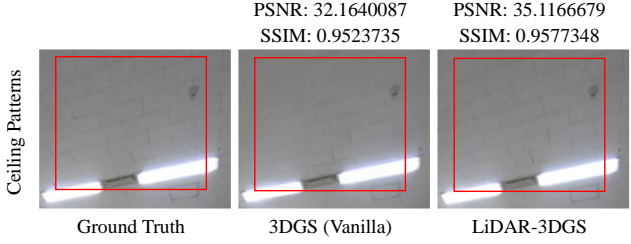
**Fig. 14.** Qualitative assessment of ceiling patten model.

In Fig. 15, The vanilla 3DGS model failed to capture the brick-like patterns in the ceiling, while the LiDAR-3DGS model successfully reflected these small details. We can see that the LiDAR point cloud shines especially when areas that have few or no SfM feature points are modeled. Because walls, floors, and objects are often homogenous in texture and therefore, lacking distinctive features for feature matching, it creates a void or a very sparse point cloud that 3D Gaussians cannot properly form. However, addition of LiDAR inputs provides foundational points for Gaussian splat to form in places where SfM points are too sparse.

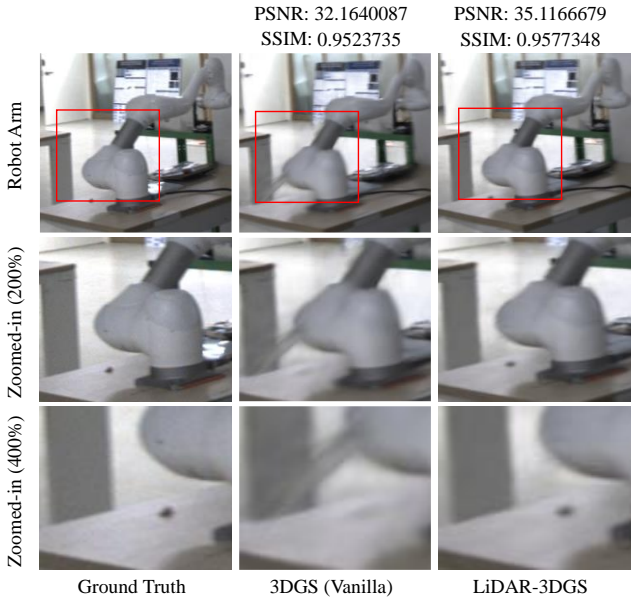
**Fig. 15.** Qualitative assessment of robot arm model.

The vanilla 3DGS model did not capture the black marker cap placed on the table beside the robot arm. It also has a 'floater' near the robot arm. Floaters are the floating glitches or errors that are created due to a Gaussian splat forming over a noise. We can see that the LiDAR-3DGS model accurately reflected the black marker cap on the table and corrected the floater. This improvement is attributed to the LiDAR points supporting the SfM features, which allowed for a precise delineation between the robot arm's geometry and the background.

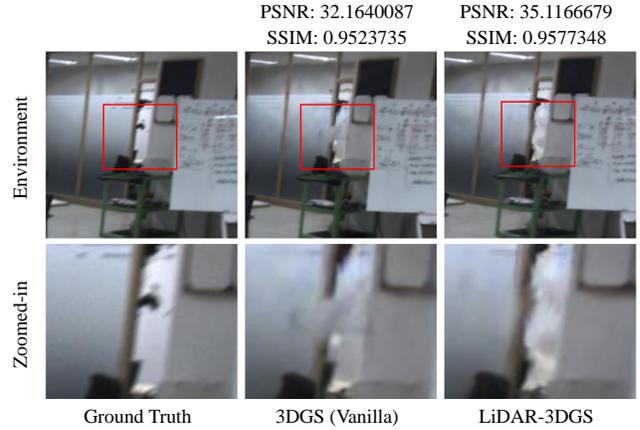
**Fig. 16.** Qualitative assessment of environment model.

The vanilla 3DGS model was unable to accurately model the entrance to the room. While the LiDAR-3DGS model did not perfectly model the entrance either, we can see an improvement. This discrepancy may be due to a mistake in data collection, such as having only a few images taken from that area to reconstruct a 3D model accurately, or the LiDAR sensor's inability to sense the entrance properly due to gaps in the laser scanning angles.

### D. Quantitative Evaluation

In the assessment of the LiDAR-3DGS models, two key metrics were utilized: PSNR and SSIM. We averaged results for every LiDAR density and it turns out that increasing trend of quality does not continue forever. The trend peaked at n=10 and decreased when too much LiDAR point clouds were added.

TABLE VIII
PSNR AND SSIM ANALYSIS

| SH/density | SfM | n = 1 | n = 5 | n = 10 | n = 20 |
|---|---|---|---|---|---|
| Average PSNR | 32.4545083 | 34.5352233 | 34.5774313 | **34.7469431** | 34.6386363 |
| Increase in PSNR | - | 2.080715 (+6.411%) | 2.122923 (+6.541%) | **2.292435 (+7.064%)** | 2.18428 (+6.730%) |
| Average SSIM | 0.952668 | 0.9566172 | 0.957312 | **0.958066** | 0.957915 |
| Increase in SSIM | - | 0.0039596 (+0.415%) | 0.004644 (+0.488%) | **0.005398 (+0.564%)** | 0.005248 (+0.548%) |

To benchmark the results against other methods, training outcomes were analyzed using the Blender dataset benchmark provided by NeRF and Mip-Splatting [28]. Although this is not a direct one-to-one comparison, it enables the identification of trends in the relative improvements of PSNR and SSIM when compared to the baseline provided by vanilla 3DGS.

TABLE IX
BENCHMARK BETWEEN VARIOUS DERIVATIVE WORKS

|  | PSNR | Relative Increase in % | SSIM | Relative Increase in % |
|---|---|---|---|---|
| 3DGS [2] | 29.77 | - | 0.960 | - |
| NeRF [1] | 31.23 | 4.904% | 0.958 | -0.208% |
| Mip-Splatting [28] | 34.56 | 16.09% | 0.979 | 1.979% |
| MipNeRF [29] | 34.51 | 15.92% | 0.973 | 1.354% |
| Plenoxels [30] | 30.34 | 1.915% | 0.955 | -0.521% |
| TensorRF [31] | 30.60 | 2.788% | 0.956 | -0.417% |
| Instant-NGP [32] | 31.20 | 4.803% | 0.959 | -0.102% |
| Tri-MipRF [33] | 34.36 | 15.42% | 0.974 | 1.458% |
| **LiDAR-3DGS (ours)** | - | **7.064%** | - | **0.564%** |

It should be noted that LiDAR-3DGS introduces an alternative approach to augment 3DGS inputs with LiDAR for integration into the unmodified 3DGS framework. *Hence, LiDAR-3DGS can be supplementary to various adaptations of the original paper*, including Mip-Splatting and other derivative works.

Following key findings emerged from the results:

- Significant improvement in PSNR shown in Table VIII with the addition of LiDAR data indicates a considerable reduction in overall error, and better geometric precision in the 3D models. Floating errors, commonly referred to as 'Floaters' in the community, were also significantly reduced in the final model.
- The highest increase in PSNR was observed with a LiDAR density of n=10 at 7.064%. However, this trend does not continue with higher densities, indicating that further increasing LiDAR density yields diminishing increases in PSNR values, making it less rewarding considering the increase in training time.
- A modest increase in SSIM of 0.564% with a LiDAR density of n = 10 shows that while LiDAR data primarily enhances geometric accuracy, it also contributes to improving perceptual aspects such as texture and color accuracy.
- As seen in Table IX Benchmarks, other works have SSIM improvements that are less than 1%, with some even showing decreases. Therefore, a 0.564% improvement is a decent result, especially considering that the training algorithm itself remains unmodified.
- The trend in PSNR improvements with increasing SH level increments varies, suggesting a dataset-specific response to changes in SH levels. This inconsistency indicates that the optimal SH level increment for maximizing PSNR can differ significantly depending on the characteristics of the dataset.
- The PSNR improvement seems to peak around an SH level increment of 1500. This can be seen as extended training periods at lower SH levels allow for a more detailed and precise foundation of Gaussian Splats, which improves the initial model accuracy.
- The lack of significant trend in SSIM with increasing SH level increments suggests that the level increments do not affect the SSIM assessment criteria. It could also be that peak SSIM converges to one point when trained for higher iterations, such as 30k iterations regardless of change in SH level increments since SH level maxes out in early iterations when compared to total numbers iterations.
- The integration of LiDAR data with SfM before training 3DGS significantly enhances the overall quality of the models. As shown in Fig. 14 to Fig. 17, LiDAR-3DGS effectively corrects or greatly reduces numerous modeling errors.
- Using this multimodal approach, the models were capable of capturing detailed components such as bolts and holes that were not captured in the vanilla 3DGS model which were presumably unmodeled due to the intrinsic problem of image-based feature matching. We can see that these intrinsic problems were compensated through the use of LiDAR point cloud.

*E. Limitations and Drawbacks*

The quality of the study is heavily affected by the quality of the LiDAR. LiDAR has range limits and gaps between its shooting angles. While camera data can collect feature points from far away as long as overlaps exist, LiDAR data is limited by the operator's physical position. If LiDAR is out of range, data cannot be collected. Even within range, gaps between LiDAR shooting angles may cause small details to be missed. Camera data must then be relied upon to match those features that are smaller than LiDAR's resolution. The Ouster OS0-32 LiDAR used in this study is a commercial-grade model with the lowest vertical resolution in its lineup. Consequently, the results of the paper are minimal. Better results can be achieved with higher-grade LiDAR systems.

TABLE X
TRAINING TIME ANALYSIS

| Density | SfM | n = 1 | n = 5 | n = 10 | n = 20 |
|---|---|---|---|---|---|
| Training Time (30k iteration) | 19m 51s | 20m 30s | 22m 17s | 23m 3s | 23m 44s |
| Increase | - | +39s (+3.275%) | +2m 26s (+12.26%) | +3m 12s (+16.12%) | +3m 53s (+19.56%) |

Training Time increases as more LiDAR data is increased. There is an increasing trend in training time as more LiDAR data is introduced. For n=1, which is the minimal LiDAR data we can introduce to the system, the increase in time was not dramatic (+3.275%) and still yielded significant increases in PSNR and SSIM (+6.411% and +0.415%, respectively). However, with denser LiDAR data such as n=20, the training time increased by 19.56%, which may significantly impact the training process as the dataset becomes larger. In applications where training resources and time are prioritized, introducing minimal LiDAR data can significantly increase quality without excessively increasing training time.

V. CONCLUSION

This paper presented and validated the multimodal approach to 3DGS by reinforcing LiDAR point cloud into 3DGS input point cloud. Several key findings were shown through the results such as enhanced detail reflection shown in the quality assessment and also supported by significant PSNR and SSIM improvement indicated clear in the performance metrics data. Models created using our method had enhanced geometric precision and was able to reflect small objects such as marker caps (Fig. 16) that were not reflected in the vanilla 3DGS. It has a shown in great reduction of noise (floaters) present in the model. As seen in Fig. 17, LiDAR-3DGS captured geometry missed by the vanilla 3DGS. This is likely because, during data collection, the object was not captured clearly or passed by too quickly for many features to be captured by the camera. However, the LiDAR point cloud was able to help form Gaussians by reinforming image-based features.

The highest average PSNR increase (7.064%) was observed with addition of LiDAR density of n=10, though further increases in LiDAR density did not continue to improve PSNR. The modest increase in SSIM (0.564%) at LiDAR density of n=10 has shown improved perceptual quality of the models. Varying increment steps for SH level gave varying results. It was initially speculated that, extended training periods at lower SH levels allow more foundational Gaussian splats to be formed. However, for maximum PSNR peaked at 1500 on our dataset. However, this may be dataset-specific to this study. SSIM remained largely unaffected by changes in SH level increments. It was speculated that having a higher SH level increment would result in longer times for lower SH Gaussian splats to form more foundations.

We can see that this multimodal input approach is a simple but powerful method of improving 3D Gaussian Splatting without having to modify the underlying algorithm. We hope that this paper reminds the importance of integration of LiDAR in Radiance Field Rendering.

REFERENCES

[1] B. Mildenhall, P. P. Srinivasan, M. Tancik, J. T. Barron, R. Ramamoorthi, and R. Ng, "NeRF: Representing Scenes as Neural Radiance Fields for View Synthesis," in Proceedings of the European Conference on Computer Vision (ECCV), 2020, pp. 405-421.

[2] B. Kerbl, G. Kopanas, T. Leimkühler, and G. Drettakis, "3D Gaussian Splatting for Real-Time Radiance Fields Rendering," ACM Transactions on Graphics, vol. 42, no. 4, pp. 1-22, 2023.

[3] J. Choi, C. M. Yeum, S. J. Dyke, and M. Jahanshahi, "Computer-aided Approach for Rapid Post-event Visual Evaluation of a Building Facade," Sensors, vol. 18, no. 9, p. 3017, 2018.

[4] J. Choi, J. A. Park, S. J. Dyke, C. M. Yeum, X. Liu, A. Lenjani, and I. Bilionis, "Similarity Learning to Enable Building Searches in Post-event Image Data," Computer-Aided Civil and Infrastructure Engineering, 2021.

[5] E. Brock, C. Huang, D. Wu and Y. Liang, "Lidar-Based Real-Time Mapping for Digital Twin Development," 2021 IEEE International Conference on Multimedia and Expo (ICME), Shenzhen, China, 2021, pp. 1-6, doi: 10.1109/ICME51207.2021.9428337.

[6] J. Choi and S. J. Dyke, "CrowdLIM: Crowdsourcing to enable lifecycle infrastructure management," Computers in Industry, vol. 115, p. 103185, 2020.

[7] C. Weitkamp, Lidar: Range-Resolved Optical Remote Sensing of the Atmosphere, Springer, 2005.

[8] W. Zhen, Y. Hu, H. Yu and S. Scherer, "LiDAR-enhanced Structure-from-Motion," in 2020 IEEE International Conference on Robotics and Automation (ICRA), Paris, France, 2020, pp. 6773-6779, doi: 10.1109/ICRA40945.2020.9197030.

[9] T. Tao, et al., "LiDAR-NeRF: Novel lidar view synthesis via neural radiance fields," arXiv preprint, 2023. Available: https://arxiv.org/abs/2304.10406.

[10] J. Bickford and C. Meering, "LiDAR for Autonomous System Design: Object Classification or Object Detection?" Analog Dialogue, Analog Devices, Jul. 2019. [Online]. Available: https://www.analog.com/en/resources/analog-dialogue/articles/lidar-for-autonomous-system-design-object-classification-or-object-detection.html.

[11] B. Xiong, Z. Li, and Z. Li, "GauU-Scene: A Scene Reconstruction Benchmark on Large Scale 3D Reconstruction Dataset Using Gaussian Splatting," arXiv preprint, 2024. Available: https://arxiv.org/abs/2401.14032.

[12] J. Lin, C. Zheng, W. Xu, and F. Zhang, "R2 live: A robust, real-time, lidar-inertial-visual tightly-coupled state estimator and mapping," IEEE Robotics and Automation Letters, vol. 6, no.4, pp. 7469–7476, 2021.

[13] T. Shan, B. Englot, C. Ratti, and D. Rus, "Lvi-sam: Tightly-coupled lidar-visual-inertial odometry via smoothing and mapping," arXiv preprint arXiv:2104.10831, 2021.

[14] X. Zuo, P. Geneva, W. Lee, Y. Liu, and G. Huang, "Lic-fusion: Lidarinertial-camera odometry," in 2019 IEEE/RSJ International Conference on Intelligent Robots and Systems (IROS). IEEE, 2019, pp. 5848–5854.

[15] X. Zuo, Y. Yang, J. Lv, Y. Liu, G. Huang, and M. Pollefeys, "Lic-fusion 2.0: Lidar-inertial camera odometry with sliding-window plane-feature tracking," in IROS 2020, 2020.

[16] J. Lin and F. Zhang, "R3LIVE: A Robust, Real-time, RGB-colored, LiDAR-Inertial Visual tightly-coupled State Estimation and mapping package," arXiv preprint, 2021. Available: https://arxiv.org/abs/2109.07982.

[17] J. Lin and F. Zhang, "R3LIVE: A Robust, Real-time, RGB-colored, LiDAR-Inertial-Visual tightly-coupled state Estimation and mapping package," in 2022 International Conference on Robotics and Automation (ICRA), 2022.

[18] ROS Wiki, "Camera Calibration," 2023. [Online]. Available: https://wiki.ros.org/camera_calibration.

[19] K. Koide, "Direct Visual-LiDAR Calibration," GitHub Repository, 2023. [Online]. Available: https://github.com/koide3/direct_visual_lidar_calibration.

[20] Ouster, Inc., "OS0 Ultra-Wide View High-Resolution Imaging Lidar," San Francisco, CA. [Online]. Available: https://ouster.com/products/hardware/os0-lidar-sensor.

[21] H. Lim, H. Chang, and J. B. Choi, "ChromaFilter: Color-Based LiDAR Filter for Real-Time Feature Extraction and Optimization," presented at the Prognostics and Health Management Korea 2024 Annual Conference (KSPHM2024), Jun. 2024.

[22] E. Rublee, V. Rabaud, K. Konolige, and G. Bradski, "ORB: An efficient alternative to SIFT or SURF," in Proceedings of the IEEE International Conference on Computer Vision (ICCV), 2011, pp. 2564-2571, doi: 10.1109/ICCV.2011.6126544.

[23] E. Rosten and T. Drummond, "Machine learning for high-speed corner detection," in Proceedings of the European Conference on Computer Vision (ECCV), 2006, pp. 430-443, doi: 10.1007/11744023_34.

[24] M. Calonder, V. Lepetit, C. Strecha, and P. Fua, "BRIEF: Binary Robust Independent Elementary Features," in Proceedings of the European Conference on Computer Vision (ECCV), Heraklion, Crete, Greece, 2010, pp. 778-792, doi: 10.1007/978-3-642-15561-1_56.

[25] "Camera Models," COLMAP, [Online]. Available: https://colmap.github.io/cameras.html.

[26] "ICP - Iterative Closest Point," CloudCompare, [Online]. Available: https://www.cloudcompare.org/doc/wiki/index.php/ICP.

[27] A. Horé and D. Ziou, "Image Quality Metrics: PSNR vs. SSIM," *2010 20th International Conference on Pattern Recognition*, Istanbul, Turkey, 2010, pp. 2366-2369, doi: 10.1109/ICPR.2010.579.

[28] Z. Yu, A. Chen, B. Huang, T. Sattler, and A. Geiger, "Mip-Splatting: Alias-free 3D Gaussian Splatting," arXiv preprint, 2023. Available: https://arxiv.org/abs/2311.16493.

[29] J. T. Barron, B. Mildenhall, M. Tancik, P. Hedman, R. Martin-Brualla, and P. P. Srinivasan, "Mip-NeRF: A Multiscale Representation for Anti-Aliasing Neural Radiance Fields," in Proceedings of the IEEE/CVF International Conference on Computer Vision (ICCV), 2021.

[30] S. Fridovich-Keil, A. Yu, M. Tancik, Q. Chen, B. Recht, and A. Kanazawa, "Plenoxels: Radiance Fields without Neural Networks," in Proceedings of the IEEE/CVF Conference on Computer Vision and Pattern Recognition (CVPR), 2022.

[31] A. Chen, Z. Xu, A. Geiger, J. Yu, and H. Su, "Tensorf: Tensorial Radiance Fields," arXiv preprint, 2022. Available: https://arxiv.org/abs/2206.02776.

[32] T. Müller, A. Evans, C. Schied, and A. Keller, "Instant Neural Graphics Primitives with a Multiresolution Hash Encoding," ACM Transactions on Graphics (TOG), vol. 41, no. 4, pp. 1-15, 2022.

[33] W. Hu, Y. Wang, L. Ma, B. Yang, L. Gao, X. Liu, and Y. Ma, "Tri-MipRF: Tri-Mip Representation for Efficient Anti-Aliasing Neural Radiance Fields," in Proceedings of the IEEE/CVF International Conference on Computer Vision (ICCV), 2023.